\documentclass[sigconf,authorversion,screen]{acmart}

\AtBeginDocument{%
  \providecommand\BibTeX{{%
    \normalfont B\kern-0.5em{\scshape i\kern-0.25em b}\kern-0.8em\TeX}}}

\copyrightyear{2023} 
\acmYear{2023} 
\setcopyright{acmlicensed}\acmConference[KDD '23]{Proceedings of the 29th ACM SIGKDD Conference on Knowledge Discovery and Data Mining}{August 6--10, 2023}{Long Beach, CA, USA}
\acmBooktitle{Proceedings of the 29th ACM SIGKDD Conference on Knowledge Discovery and Data Mining (KDD '23), August 6--10, 2023, Long Beach, CA, USA}
\acmPrice{15.00}
\acmDOI{10.1145/3580305.3599438}
\acmISBN{979-8-4007-0103-0/23/08}

\usepackage{hhline}
\usepackage{bm}
\usepackage{bbm} 
\usepackage{graphicx}  
\usepackage{tabularx}
\usepackage{multirow}
\usepackage{booktabs} 
\usepackage{mathrsfs}
\usepackage{makecell}
\usepackage{amsmath}
\usepackage{amsfonts}
\usepackage[ruled,linesnumbered]{algorithm2e}  
\usepackage{algorithmic} 
\usepackage{color}
\usepackage{arydshln} 
\usepackage{setspace} 

\begin{document}

\title{MSSRNet: Manipulating Sequential Style Representation for Unsupervised Text Style Transfer}

\author{Yazheng Yang}
\orcid{0000-0003-1627-8341}
\affiliation{
  \institution{The University of Hong Kong}
  \city{Hong Kong}
  \country{China}
  }\email{yangyazh@connect.hku.hk}

\author{Zhou Zhao}
\orcid{0000-0001-6121-0384}
\affiliation{
  \institution{Zhejiang University}
  \city{Hangzhou}
  \state{Zhejiang}
  \country{China}
  }\email{zhaozhou@zju.edu.cn}

\author{Qi Liu}
\orcid{0000-0003-4608-5778}
\affiliation{
  \institution{The University of Hong Kong}
  \city{Hong Kong}
  \country{China}
  }\email{liuqi@cs.hku.hk}

\renewcommand{\shortauthors}{Yazheng Yang, Zhou Zhao, \& Qi Liu}

\begin{abstract}
Unsupervised text style transfer task aims to rewrite a text into target style while preserving its main content. 
Traditional methods rely on the use of a fixed-sized vector to regulate text style, which is difficult to accurately convey the style strength for each individual token. In fact, each token of a text contains different style intensity and makes different contribution to the overall style. 
Our proposed method addresses this issue by assigning individual style vector to each token in a text, allowing for fine-grained control and manipulation of the style strength. Additionally, an adversarial training framework integrated with teacher-student learning is introduced to enhance training stability and reduce the complexity of high-dimensional optimization. The results of our experiments demonstrate the efficacy of our method in terms of clearly improved style transfer accuracy and content preservation in both two-style transfer and multi-style transfer settings.\footnote{Code is available: https://github.com/OldBirdAZ/mssrnet\_style\_transfer}
\end{abstract}

\ccsdesc[500]{Computing methodologies~Natural language generation}
\ccsdesc[300]{Computing methodologies~Machine learning algorithms}
\ccsdesc[500]{Theory of computation~Unsupervised learning and clustering}

\keywords{Text Style Transfer, Unsupervised Learning, Adversarial Training, Teacher-Student Learning, Natural Language Processing}


\maketitle

\begin{figure}[htp]
    \includegraphics[width=0.5 \textwidth]{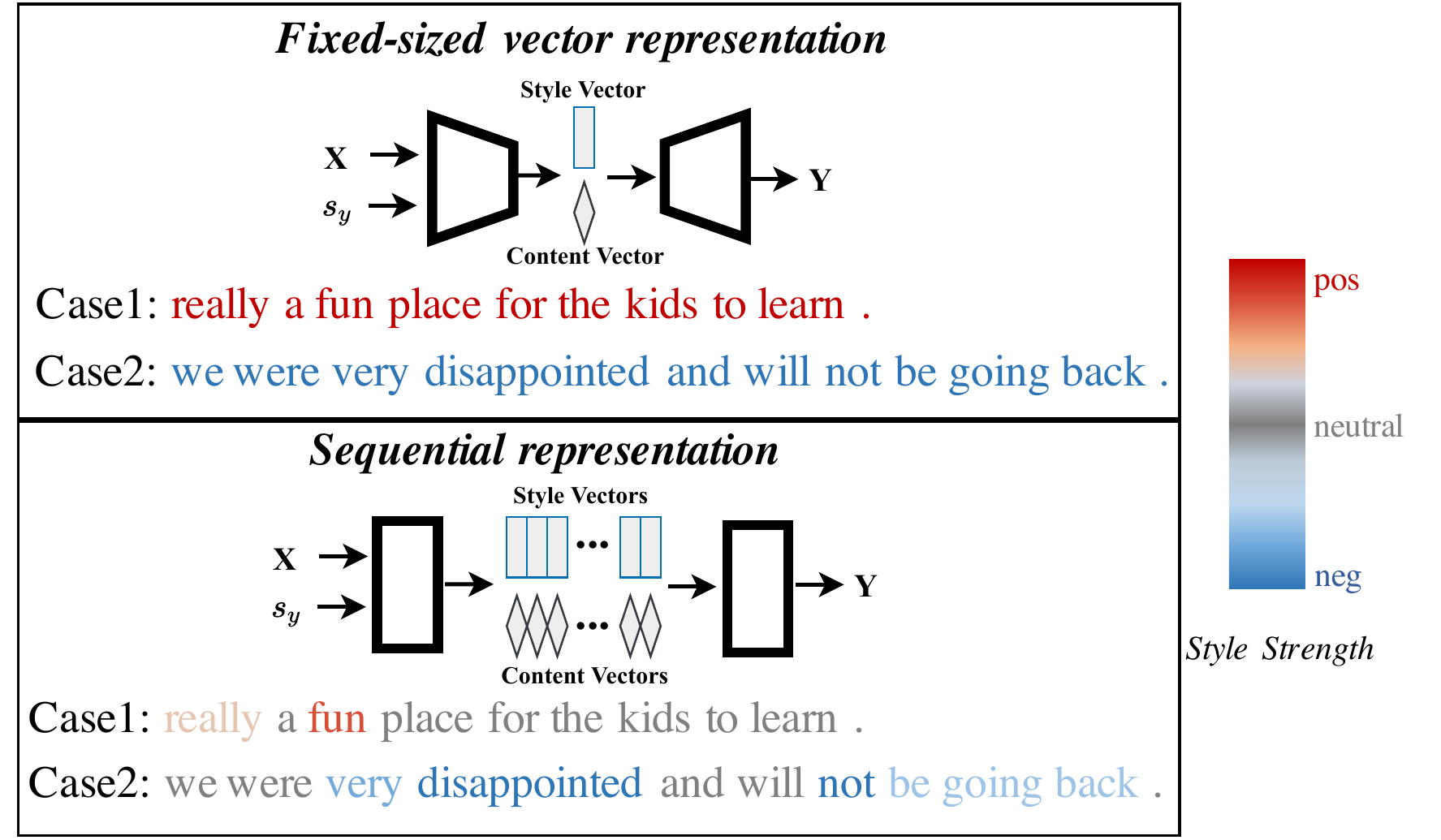}
    \caption{\label{fig_intro_example} Comparison between fixed-sized vector and sequential representations with two examples selected from Yelp dataset. Style strength differs from token to token contributing to the overall style. 
    The fixed-sized vector applies the same style information to all words forming coarse-grained control and is hard to convey the differences among tokens, while manipulating style information for each token individually can distinguish such differences leading to fine-grained management.
    }
\end{figure}
\section{Introduction}
Text style transfer is a practical task in the field of Natural Language Processing~(NLP) that teaches machine to rewrite a text into a different style (e.g. sentiment, author writing style) while retaining its content. Due to the scarcity of parallel datasets with varied styles, we focus on the unsupervised text style transfer in this work.

The dominant approaches \cite{hu2017toward,shen2017style,fu2018style,john-etal-2019-disentangled} in unsupervised text style transfer typically involve disentangling the style representation from content representation, that both are compressed into a fixed-sized vector separately. 
However, this approach has some limitations. The fixed-sized style representation is not capable of capturing the subtle differences in style intensity among the different tokens of a text. As a result, it becomes difficult to achieve fine-grained manipulation of the style. Figure~\ref{fig_intro_example} shows two examples whose overall sentiments are positive and negative respectively. Considering its context, the first example demonstrates that some tokens (e.g. ``really'', ``fun'') are essential to express positive sentiment while the others are neutral. Actually, not all tokens are of the same style strength in most cases. Even though the fixed-sized vector is adequate for the need of controlling the coarse-grained style, it is hard to accomplish the fine-grained manipulation. Empirically, the model needs a large number of efforts to distinguish style-independent contents from style-dependent information because it uses the single style vector for all tokens. The results of low content preservation shown by prior work that adopted vector representation also indicate the existence of this problem. 
Additionally, in order to incorporate with style vector, the content information is also typically squashed into a single fixed-size vector that often loses significant semantic and positional information, preventing the decoder from referring to the hidden states of the input sequence during the decoding stage. This limits the content preservation of the transferred text. 
The ability to refer to the input sequence has been shown to be beneficial in Neural Machine Translation (NMT) \cite{bahdanau2014neural} and has the potential to improve content preservation in text style transfer as well. 
The experimental results of some works \cite{lample2018multiple,dai-etal-2019-style} show that referring to the source sequence will improve the content preservation.

\begin{figure*}[htp]
\begin{center} 
    \includegraphics[width=1.0\linewidth]{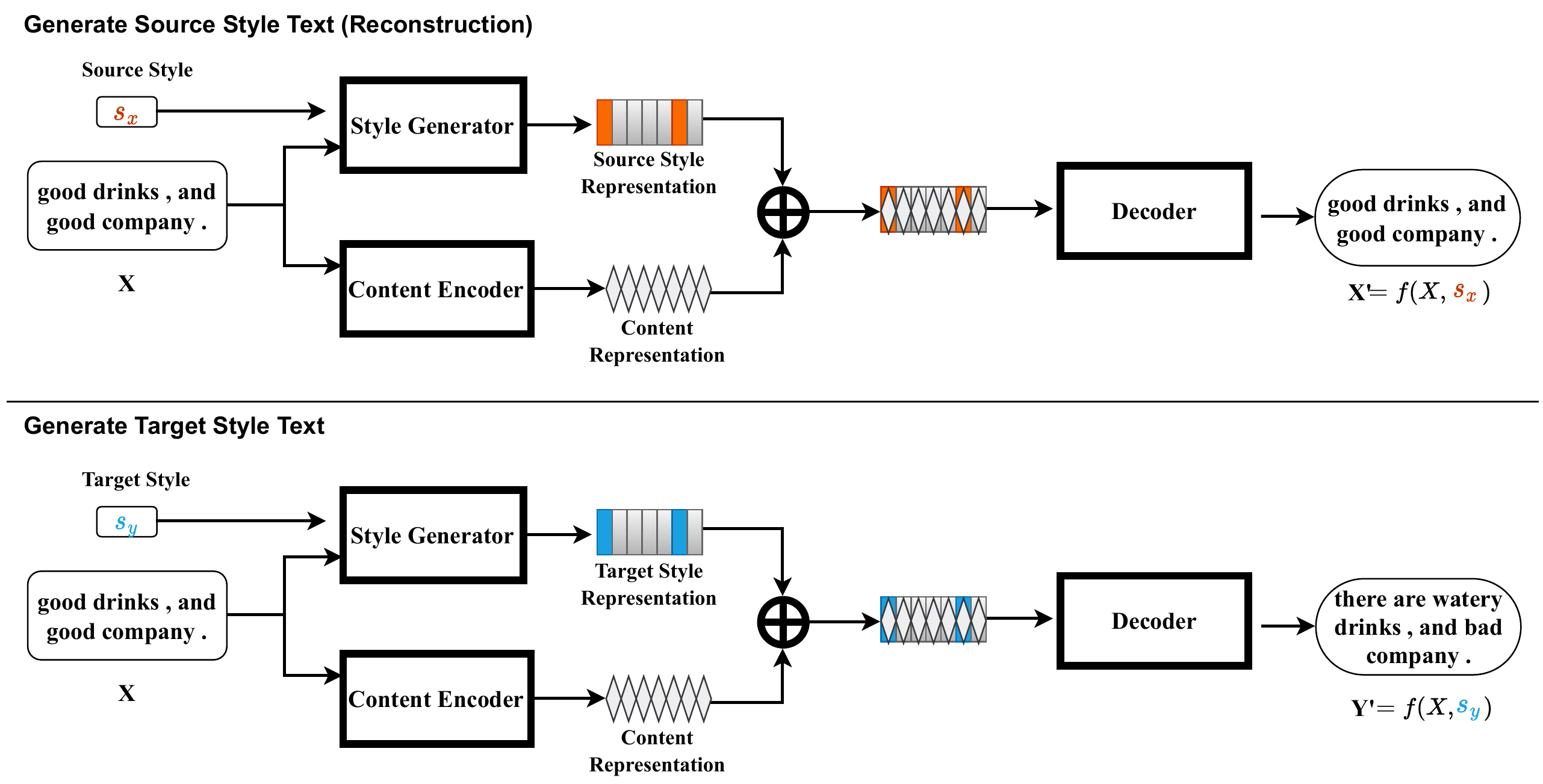} 
    \caption{\label{fig_overall}The overall architecture of our model which consists of the style generator, content encoder and the decoder. At inference time, our model takes $\mathbf{X}$ and target style indicator $s_y$ as input, and then generates $\mathbf{Y}^{'}$ as output. Here we also depict the self-reconstruction procedure $\mathbf{X}^{'}=f(X, s_x)$. Note that our model is designed for the scenario of multiple styles, we just show the case of two styles for simplification. We can set the target style indicator $s_y$ to be any style index for multi-style transfer. 
    }
\end{center}
\end{figure*}
 
To address these limitations, we propose the use of \textit{sequential style representation}, in which a separate vector is generated for each token of the source text. This allows for precise control of the style strength of each token. Our proposed model, called the \textit{Manipulating Sequential Style Representation Network} (MSSRNet), controls this representation space. The advantage of this approach is that it enables the model to easily refer to the sequence (combination of the content and target style representation) during the decoding stage and utilize attention mechanisms. This leads to improved content preservation in the transferred text. 
Specifically, our MSSRNet consists of a \textit{Style Generator}, a \textit{Content Encoder} and the \textit{Decoder}. Given source text $\mathit{X}$ and target style label $\mathit{s_y}$, the style generator learns to produce sequential style representation $\textbf{\emph{S}}_y$. The content encoder generates content representation of $\mathit{X}$. After combining style representation with content representation, the decoder takes this combination as input and generates target text. 
Similar to previous methods, we adopt the training framework of Generative Adversarial Network~(GAN)~\cite{goodfellow2014generative,arjovsky2017wasserstein} in this unsupervised task.

However, manipulating sequential style representation is a more complex task compared to controlling a fixed-sized vector representation due to its higher dimensionality. The challenge lies in effectively navigating and adjusting the style representation for each individual token in the sequence. To address this, we introduce a novel framework that uses dual discriminators: a style discriminator and a text discriminator. The former controls the quality of the generated style representation, while the latter ensures that the generated text is in the target style. Additionally, to provide stable feedback signals to the style generator and enhance training stability, we incorporate Teacher-Student learning~\cite{bucilu2006model,hinton2015distilling} into the adversarial training framework. The hidden states of a text, encoded by the style classifier, represent its stylistic information. As such, the style classifier acts as a teacher model, guiding the style generator to produce the desired style representation and penalizing it when it fails to generate the target style representation. Such supervised complementation information brings stability to the interactive optimization~\cite{pan2020ami} of the GAN training framework.

Our method has been extensively evaluated on benchmark datasets, and the experimental results demonstrate its superiority over existing approaches. The fine-grained manipulation of sequential style representation allows for better control of text style, leading to improved style transfer accuracy and better preservation of content compared to previous methods in both two-style and multi-style transfer tasks.

\section{Related Work} 
Most prior work \cite{hu2017toward,shen2017style,li2018delete,xu-etal-2018-unpaired,dai-etal-2019-style} focused on disentangling the style vector from the content vector. 
Hu~\shortcite{hu2017toward} proposed an approach to combine Variational Auto-Encoders (VAEs) \cite{kingma2013auto} and holistic attribute discriminators to control the style vector. 
Shen~\shortcite{shen2017style} proposed a cross-aligned method to separate the content vector from style information. 
They assumed that different text corpora shares the same latent content distribution and can be transferred to other style through refining the alignment of latent representation. Through aligning among different corpora, their model is trained to learns the style vector for each corpus.
Fu~\shortcite{fu2018style} also explored using the architecture that consists of the style embedding and single decoder. The decoder takes the content vector along with the style embedding as input.
To make sure the content vector preserves the main information of the text, Prabhumoye~\shortcite{prabhumoye2018style} explored using back-translation and adopted the multi-decoder architecture. 
Xu~\shortcite{xu-etal-2018-unpaired} added the constraint of cycle consistency into the back-translation with reinforcement learning \cite{williams1992simple} to encourage the model to preserve much more content.
John~\shortcite{john-etal-2019-disentangled} proposed to disentangle the content and the style vector through multi-task learning along with the adversarial training. 
However, due to the limited representation ability of fixed-sized vector, these approaches mentioned above share similar characteristics: they transfer text style with low content preservation, even though some of them achieve promising transfer accuracy, because a single vector is appropriate to represent the overall polarity of text style but it drops a large amount of content information.

In contrast to disentanglement-based methods, another line of work tried to build powerful generator. 
Dai~\shortcite{dai-etal-2019-style} used the Transformer~\cite{vaswani2017attention} as their generator of GAN framework. They believed that the neural networks need no separation between the style and content representation if the decoder is strong enough to rewrite a text into its opposite style. 
Lin~\shortcite{liu-etal-2021-learning} applied the GPT-2~\cite{radford2019language} as their model, and proposed a reward function to constrain the content preservation. Lai~\shortcite{lai2021generic} proposed BARTM to utilize a large scale generic paraphrases corpus, then synthesize a parallel dataset to finetune the pretrained model, BART~\cite{lewis2019bart}. They trained two models, each in a transfer direction so that they could improve each other with iterative back-translation approach. 
Li~\shortcite{li2020dgst} also presented a Dual-Generator Network without employing GAN training. One of the shortcomings of these methods is that they need to build a model for each transfer direction, and it's tedious to extend for multi-style scenario. 
Other approaches turned to detecting and replacing the ``pivot words'' that represent the style of the text. 
Li~\shortcite{li2018delete} proposed a pipeline to delete, retrieve and then generate the target style text, which may be trivial and hard to train. Similarly, Xiao~\shortcite{xiao2021transductive} further utilized a retriever that selects top K content-similar sentences from target style corpora to select pivot words.

Early applications of teacher-student learning usually focused on model compression or knowledge distillation \cite{bucilu2006model,hinton2015distilling}, where a small student model is trained to mimic the cumbersome teacher model or ensemble of models. The student model usually shares similar architecture with the teacher model.
In a more general perspective, the student learns specific knowledge from the teacher model or mimic some manners of the teacher model without requirement of sharing similar model structure. 
The teacher-student learning has been applied to several tasks, such as dialogue generation \cite{wang-etal-2018-teacher,peng2019teacher}, and neural machine translation \cite{chen-etal-2017-teacher,tan2019multilingual,pan2020bidecoder},  Named Entity Recognition
(NER) \cite{yirmibecsouglu2020ermi,wu2020single}, etc.
In this work, in order to improve the training stability as well as reduce the optimization difficulty, we leverage the teacher-student learning, so that our Style Generator module learns the knowledge about how to generate desired style representation from a text style classifier. 

\section{Approach}
In this section, we will give a brief description of the text style transfer task first, and then we will introduce the architecture of our model. Finally, we will present our training strategy.

Given a text $ \rm X=\{x_{1}, x_{2}, ..., x_{n}\}$ of style $\rm s_x$, the goal is to train the model $ f $ dedicating to generating a new text $ {\rm Y}=\{y_{1}, y_{2}, ..., y_{m}\} = f ({\rm X}, {\rm s_y}) $ with the target style $\rm s_y$, where n, m is the length of ${\rm X}$ and ${\rm Y}$ respectively. 

\subsection{Model Architecture}
As shown in Figure \ref{fig_overall}, our style transfer model, MSSRNet, consists of the style generator, the content encoder and the decoder. Besides, we also apply the style discriminator and text discriminator to conduct the adversarial training as well as employ the classifier as teacher model. We build our model based on modules of the well-known Transformer, but there still exist some important points.

\textbf{Style Generator} takes the encoder of Transformer as backbone. Given the source text ${\rm X}$ and target style indicator $\rm s_y$, the Style Generator produces sequential style representation:
\begin{equation}
\begin{aligned}
\{ \mathbf{h}_s, \mathbf{h}_1, ..., \mathbf{h}_n \} = {\rm Enc} ([Emb_s(s_y);Emb(X)])
\end{aligned}
\end{equation}
where $Emb_s(.)$ denotes the style embedding, $Emb$ denotes the word embedding along with positional embedding, ${\rm Enc}$ represents the encoder of Transformer, n is the length of X, and ``;'' means concatenation. Here $Emb_s(s_y)$ tells the model about which style is going to be generated in terms of current text. Finally, we discard $\mathbf{h}_s$ and take $ \textbf{\emph{S}}_y=\{ \mathbf{h}_1, \mathbf{h}_2, ..., \mathbf{h}_n \} $ as the sequential style representation of X for style $\rm s_y$ since $\mathbf{h}_1$ corresponds to the first token of X.

\textbf{Content Encoder} is also build upon Transformer encoder:
\begin{equation}
\begin{aligned}
\textbf{\emph{C}} = {\rm Enc} (Emb(X))
\end{aligned}
\end{equation}
The combination of $\textbf{\emph{S}}_y$ and $\textbf{\emph{C}}$ is taken as input of the Decoder:
\begin{equation}
\begin{aligned}
\textbf{\emph{H}} = \textbf{W}_c^{\rm T} \textbf{\emph{C}} + \textbf{W}_s^{\rm T} \textbf{\emph{S}}_y
\end{aligned}
\end{equation}
where $\textbf{W}_c^{\rm T}$ and $\textbf{W}_s^{\rm T}$ are trainable variables. 
With such setting, we can fuse the content and target style information under a new semantic space so that the decoder is convenient to refer to it at decoding stage. Compared with style information, the semantic information of content generally is more complex and difficult to store and represent so that it needs higher dimension space. In practice, we can adjust the parameter of style representation according to the difficulty of modeling style information.

\textbf{Decoder}
We adopt Transformer decoder as the decoding module of our model:
\begin{equation}
\begin{aligned}
\{y_{1}, y_{2}, ..., y_{m}\} = {\rm Dec} (\textbf{\emph{H}})
\end{aligned}
\end{equation}
where m is the length of $\rm Y$.
However, when optimizing the decoder with adversarial training, a difficulty faced is the intractability in calculating the gradients because of the discreteness of the generated text $Y$. In this case, we compute the embedding with the vocabulary distribution $\mathcal{Y}$ of $Y$, and feed $Emb(Y)$ to the text discriminator:
\begin{equation}
\begin{aligned}
Emb(Y) = \mathcal{Y} \textbf{W}_e
\end{aligned}
\end{equation}
where $\mathcal{Y} \in [0, 1]^{m \times T}$, $T$ is the vocabulary size. $\textbf{W}_e \in \mathbb{R}^{T \times Q}$ is the embedding weight matrix, here $Q$ represents the embedding size.

\subsection{Teacher Model}
The teacher Model is a text style classifier, and takes Transformer encoder as basic component:
\begin{equation}
\label{eq_classifier_states}
\begin{aligned}
\textbf{\emph{S}}_x = {\rm Enc} (Emb(X))
\end{aligned}
\end{equation}
where $\textbf{\emph{S}}_x = \{ \mathbf{s}_1, \mathbf{s}_2, ..., \mathbf{s}_n \} $ is the sequence of style vectors for tokens from $\rm X$, n is the length of $\rm X$.
In this work, we make use of the classifier as the teacher model which requires to back-propagate the gradient to the style transfer model.
We compute the representation of the whole text by the weighted average of the hidden states $\textbf{\emph{S}}_x$ of all words:
\begin{equation}
\label{eq_meanpooling}
\begin{aligned}
\textbf{\emph{v}}_x = \sum_i \frac{{\rm exp}(\mathbf{v}_1^{\top} \mathbf{s}_i)}{\sum_{i'}{\rm exp}(\mathbf{v}_1^{\top} \mathbf{s}_{i'})}\mathbf{s}_i
\end{aligned}
\end{equation}
where $\mathbf{v}_1$ is trainable vector.
%
Such average over $\textbf{\emph{S}}_x$ across the sequence length dimension will encourage $\mathbf{s}_i$ to contain more stylistic information when token $x_i$ is key word for the overall style, otherwise no/little style information while $x_i$ is neutral.
$\textbf{\emph{v}}_x$ is further passed into a linear layer followed with softmax to make style classification:
\begin{equation}
\label{eq_cls_probs}
\begin{aligned}
\textbf{\emph{d}}_x = softmax( \textbf{W}_{tm}^{\rm T} \textbf{\emph{v}}_x + \textbf{b}_{tm} )
\end{aligned}
\end{equation}
where $\textbf{W}_{tm}$ and $\textbf{b}_{tm}$ are learnable parameters, $\textbf{\emph{d}}_x \in \mathbb{R}^K $, $K$ is the number of styles.

\subsection{Discriminators}
\textbf{Style Discriminator}
Given sequential style representation $\textbf{\emph{S}}$, the style discriminator aims to judge its style $\rm s$.
\begin{equation}
\begin{aligned}
\textbf{\emph{v}}_{sd} = {\rm Avg}_{dim}(MLP(\textbf{\emph{S}}))
\end{aligned}
\end{equation}
where ${\rm Avg}_{dim}$ denotes the weighted average across the sequence length the same as Eq.(\ref{eq_meanpooling}) but uses different parameter, $MLP$ represents the Multi-Layer Perceptron layer.
$\textbf{\emph{v}}_{sd}$ is further fed into a linear layer:
\begin{equation}
\begin{aligned}
\textbf{\emph{d}}_{sd} = \textbf{W}_{sd}^{\rm T} \textbf{\emph{v}}_{sd} + \textbf{b}_{sd}
\end{aligned}
\end{equation}
where $\textbf{W}_{sd}$ and $\textbf{b}_{sd}$ are trainable parameters. $\textbf{\emph{d}}_{sd} \in \mathbb{R}^K $, $K$ is the number of styles.

\textbf{Text Discriminator}
The text discriminator shares the same network architecture as teacher model except for removing the softmax function on the output layer, because the discriminator of Wasserstein GAN~\cite{gulrajani2017improved} needs to discard the activation function. 
\begin{equation}
\begin{aligned}
\textbf{\emph{S}}_{td} = {\rm Enc} (Emb(X))\\
\textbf{\emph{v}}_{td} = \sum_i \frac{{\rm exp}(\mathbf{v}_2^{\top} \mathbf{s}_{td}^i)}{\sum_{i'}{\rm exp}(\mathbf{v}_2^{\top} \mathbf{s}_{td}^{i'})}\mathbf{s}_{td}^i \\
\textbf{\emph{d}}_{td} = \textbf{W}_{td}^{\rm T} \textbf{\emph{v}}_{td} + \textbf{b}_{td}
\end{aligned}
\end{equation}
where $\mathbf{v}_2$, $\textbf{W}_{td}$ and $\textbf{b}_{td}$ are trainable parameters. $\textbf{\emph{d}}_{td} \in \mathbb{R}^K $, $K$ is the number of styles.

\subsection{Training}
The teacher model is trained in advance before training MSSRNet. It is used to teach style generator about producing style representation properly when the target style indicator is exactly the original style indicator $s_y = s_x$. Besides, it also provides real examples for style discriminator.

\textbf{Self Reconstruction}
The style transfer model is required to generate $X \leftarrow f ({\rm X}, {\rm s_y}) $ when $s_y = s_x$. In order to prevent the model from directly copying the source text $X$ which may encourage the content representation to contain source style information, we make noise through randomly replacing some tokens of $X$ with unknown token $<$unk$>$. The loss for self reconstruction is computed as negative log-likelihood:
\begin{equation}
\label{eq_self_recons_ojb}
\begin{aligned}
\mathcal{L}_{cst}(\theta) = - log (p (X|X', s_x)) 
\end{aligned}
\end{equation}
where $X'$ means the noised text for $X$, $\theta$ is the trainable parameters of MSSRNet.

\textbf{Adversarial Training}
The style representations generated by teacher model are treated as real (or positive) examples for style $s_x$, while those $f_{\phi}(X, s_x)$ synthesized by style generator $f_{\phi}$ are fake (or negative) examples when we train style discriminator. The min-max game between the style generator and the style discriminator is that the style generator aims to produce style information to cheat the style discriminator while the style discriminator dedicates to distinguishing real examples from fake ones. In a similar manner, the text discriminator regards $X$ from dataset as positive examples and those produced by style transfer model as negative ones. The min-max optimization over the generator and discriminator $f_D$ can be formulated as:
\begin{equation}
\label{gan_opt}
\begin{aligned}
\min_{\theta} \max_{w \in \mathcal{W}} \mathbb{E}_{\textbf{z} \sim \mathbb{P}_r} [f_D(\textbf{z})] - \mathbb{E}_{\tilde{\textbf{z}} \sim \mathbb{P}_g} [f_D(\tilde{\textbf{z}})]
\end{aligned}
\end{equation}
where $\textbf{z}$ and $\tilde{\textbf{z}}$ represent the positive and negative example respectively. $\theta$ and $w$ here are parameters of the generator and the discriminator respectively. $\mathbb{P}_r$ and $\mathbb{P}_g$ represent the real distribution and the generated data distribution. 
In the GAN training framework, we obtain $\mathcal{L}_{adv}(\phi)$,$\mathcal{L}_{adv}(\theta)$,$\mathcal{L}_{adv}(\mathcal{W}_s)$ and $\mathcal{L}_{adv}(\mathcal{W}_t)$ representing the adversarial loss for style generator module, style transfer model, style discriminator and text discriminator respectively.

\textbf{Learning From Teacher Model}
For the case $s_y = s_x$, the style generator is expected to produce $\textbf{S}_y$ approaching to $\textbf{S}_x$ generated by teacher model in Eq.(\ref{eq_classifier_states}). We calculate the Mean-Squared Error (MSE) between $\textbf{S}_y$ and $\textbf{S}_x$ as its loss:
\begin{equation}
\label{eq_s_cls_teach_obj}
\begin{aligned}
\mathcal{L}_{teach}(\phi) = {\rm MSE} (\textbf{S}_y, \textbf{S}_x) 
\end{aligned}
\end{equation}
where $\phi$ denotes the trainable parameters of style generator module.

Meanwhile, in order to enforce the overall style, we feed the generated style representation $\textbf{S}_y$ into teacher model, and calculate the Cross-Entropy Loss:
\begin{equation}
\label{eq_s_cls_obj}
\begin{aligned}
\mathcal{L}_{s\_polarity}(\phi) = {\rm CELoss} (f_{TM}^{'} (\textbf{S}_y), s_y) 
\end{aligned}
\end{equation}
where $f_{TM}^{'}(.)$ represents the calculation procedure as shown in Eq.(\ref{eq_meanpooling}) and Eq.(\ref{eq_cls_probs}).
Similarly, we also penalize the generated text $Y$ with teacher model, so that it is of the target style $s_y$:
\begin{equation}
\label{eq_t_cls_obj}
\begin{aligned}
\mathcal{L}_{t\_polarity}(\theta) = {\rm CELoss} (f_{TM} (Y), s_y) 
\end{aligned}
\end{equation}
where $f_{TM} (.)$ denotes the teacher model.

Since the model is trained lacking of paralleled dataset, traditional yet practical approaches rely on GAN framework. Unfortunately, such adversarial training is well known of being delicate and unstable~\cite{arjovsky2017wasserstein}. Theoretically, providing supervised signal of optimization from the teacher model will promote its stability.

\begin{algorithm}[htbp] 
    \caption{Training Procedures} 
    \label{alg_training_proc} 
    \begin{algorithmic}[1]
        \STATE \textbf{Input:}dataset $\mathcal{D}$, training step $N$, number of styles $K$, $N_{rc}$,$N_{dr}$,$N_{adr}$ are the number of training iteration.
        \STATE \textbf{Init:} train the classifier with training set.
        \FOR{each iteration i = 1,2,...,N}
        \FOR{each j = 1,2,...,$N_{rc}$}
        \STATE train model with Eq.(\ref{eq_self_recons_ojb}$\sim$\ref{eq_t_cls_obj});
        \ENDFOR
        \FOR{each j = 1,2,...,$N_{dr}$}
        \STATE train two discriminators through minimizing $\mathcal{L}_{adv}(\mathcal{W}_s)$ and $\mathcal{L}_{adv}(\mathcal{W}_t)$;
        \ENDFOR
        \FOR{each j = 1,2,...,$N_{adr}$}
        \STATE sample $s_y$ from $K$ styles, $s_y \neq s_x$;
        \STATE $Y, \textbf{\emph{S}}_y \leftarrow f ({\rm X}, s_y)$;
        \STATE train model with adversarial feedback: minimize $\mathcal{L}_{adv}(\phi)$ and $\mathcal{L}_{adv}(\theta)$;
        \ENDFOR
        \ENDFOR 
    \end{algorithmic}
\end{algorithm}

\begin{figure}[htbp]
    \includegraphics[width=0.48 \textwidth]{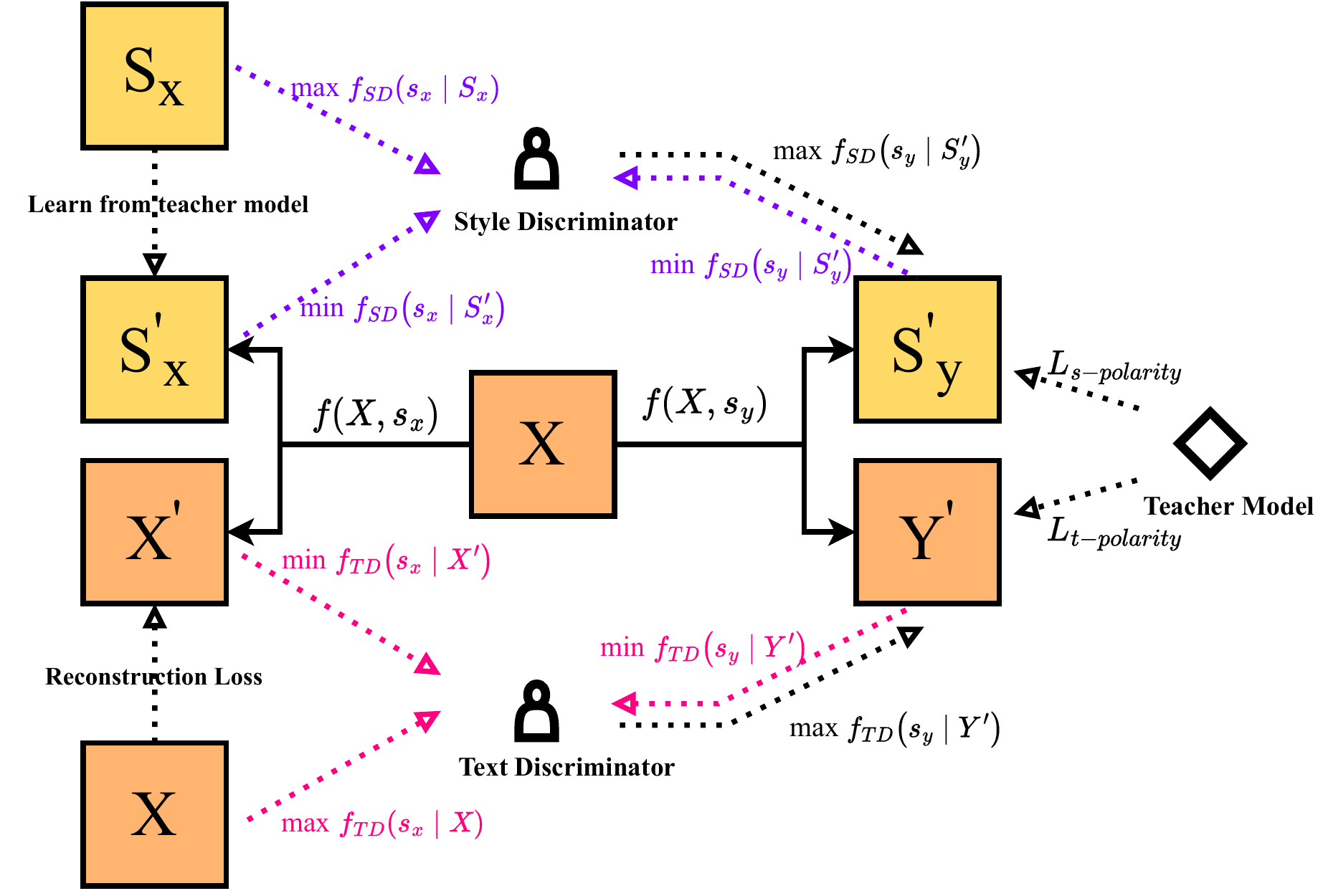}
    \caption{\label{fig_optimization_bp} Illustration of our training framework within which adversarial training and teacher-student learning are applied. The arrows with dash denote the back-propagate direction. \textbf{S}$_{\rm x}=f_{TM}(\textbf{X})$ represents the sequential style representation produced by teacher model. $f_{SD}(.)$ and $f_{TD}(.)$ denote style discriminator and text discriminator.
    }
    \vspace{-2ex}
\end{figure}

\textbf{Training Procedures}
How to conduct the training of our MSSRNet are illustrated in Algorithm~\ref{alg_training_proc}. For each iteration, we first train our model with objectives of self reconstruction and teacher-student learning as described on Eq.(\ref{eq_self_recons_ojb}$\sim$\ref{eq_t_cls_obj}) with $N_{rc}$ batches of training data. We then train discriminators with $N_{dr}$ batches of training data so that discriminators are able to learn to distinguish real examples from synthetic ones. Finally, we apply discriminators to encourage MSSRNet to generate examples that approach to real distributions.
Figure~\ref{fig_optimization_bp} demonstrates the optimization directions within our training framework. The purple arrows represent the direction of optimizing style discriminator, while those red ones represent the direction of training text discriminator. Similarly, the black dash arrows denote optimizing MSSRNet.

\section{Experiments}
\subsection{Datasets}

\textbf{Yelp} A collection of restaurants and business reviews with user ratings. For comparison, we take the preprocessed version\footnote{https://github. com/shentianxiao/language-style-transfer}, which has been widely used in previous work. 
We also use the reference set of 1000 reviews released by Li~\shortcite{li2018delete} for evaluation, where the paralleled reviews contain the same content with reverse sentiment. There are 443k/4k/1k sentences for training/validation/testing respectively.\\
\textbf{IMDb} Another common used corpus for this task, which consists of movie reviews. For fairness and reproduction, we use the preprocessed dataset\footnote{https://github.com/fastnlp/nlp-dataset} by Dai~\shortcite{dai-etal-2019-style}. The preprocessed dataset consists of 366K/4k/2k sentences for training/validation/testing respectively. \\
\textbf{Sentiment-Formality} A dataset for multiple style transfer task across different stylistic dimensions~\cite{goyal2021multi}, which are positive sentiment, negative sentiment, formal writing and informal writing styles. There are 548k/6k/6k sentences for training/validation/testing respectively.

\subsection{Metrics}
\label{metric_section}
To evaluate the performance of different models, we use 4 automatic metrics: \\
\textbf{Accuracy } The accuracy of successfully changing text into target style is measured by an extra classifier trained on each dataset. Following \citet{dai-etal-2019-style}, we use fastText \cite{joulin-etal-2017-bag} to train the classifier and achieve accuracy of 97\%, 99\% and 92\% on the test set of Yelp, IMDb and Sentiment-Formality respectively. Therefore, it can provide a quantitative way of evaluating the transfer accuracy. \\
\textbf{PPL } To measure the fluency of generated sentences, we train 5-gram language models with KenLM \cite{heafield2011kenlm} on the Yelp, IMDb and Sentiment-Formality dataset respectively. \\
\textbf{BLEU } The BLEU metric is widely used in the neural machine translation task~\cite{chen-etal-2017-teacher,tan2019multilingual}. Following prior work, we use the BLEU to measure the content consistency between transferred text and its original text in this task. \\
\textbf{r-BLEU } We compute the BLEU score between transferred text and its ground truth if it is available on the test set. We compute the r-BLEU score on Yelp and Sentiment-Formality based on their corresponding paralleled reference test sets.

\subsection{Baselines}
\textbf{DelAndRet}\footnote{https://github.com/lijuncen/ Sentiment-and-Style-Transfer} With the observation that style transfer is often accomplished by altering some attribute markers, Li etc. \shortcite{li2018delete} proposed a pipeline in which the content words are extracted by deleting those style-relevant phrases, then target style phrases are retrieved from opposite corpus, and finally a new sentence is generated conditioned on the remaining words and
those retrieved phrases. \\
\textbf{MultiDecoder}\footnote{https://github.com/fuzhenxin/text\_style\_transfer} The MultiDecoder model tries to use single encoder to take the content representation, and utilizes different decoder, one for each style, to produce text in desired style. The model is also trained within the GAN framework.  \\
\textbf{StyTrans-cond}\footnote{https://github.com/fastnlp/style-transformer} Both StyTrans-cond and StyTrans-multi are also trained with adversarial training. Dai etc.~\shortcite{dai-etal-2019-style} argued that disentanglement between content information and style information is unnecessary, since its quality is difficult to judge. They built their model based on Transformer. The discriminator of StyTrans-cond takes a sentence \textit{x} and a style indicator \textit{s} as input, and answers whether the input sentence has the corresponding style. \\
\textbf{StyTrans-multi} Different from StyTrans-cond, the discriminator of StyTrans-multi only takes a sentence \textit{x} as input, and then predicts the style of this sentence.  \\
\textbf{DGST}\footnote{https://github.com/lissomx/DGST} DGST discards adversarial training, and adopts a special noising method, called neighbourhood sampling, that is able to add noise to each sentence dynamically. There are two generators, one for each transfer direction, that are expected to interactively improve each other. Besides, in order to achieve better content preservation, the back-translation ($\textit{X}\rightarrow\textit{Y}\rightarrow\textit{X}$) is applied.  \\
\textbf{DIRR}\footnote{https://github.com/yixinL7/Direct-Style-Transfer} DIRR is based on GPT-2~\cite{radford2019language} through forming such conditional generation as a sequence completion task. They concatenated the source
sentence with a special token that serves as style indicator. Besides, they also presented a special reward function to guarantee the content similarity. \\
\textbf{BARTM}\footnote{https://github.com/laihuiyuan/Generic-resources-for-TST} BARTM is trained with multiple procedures: (1) finetune BART~\cite{lewis2019bart} on existing generic paraphrase corpora and synthetic pairs produced by leveraging a general-purpose lexical resource; (2) apply iterative back-translation to further finetune two BART models, one for each transfer direction. (3) generate synthetic pairs as paralleled dataset to train BART as final model.

\subsection{Implementation Details}
The word embedding size, positional embedding size and the hidden size of other modules are all set as 256. The feed-forward size sets to be 512. The encoder and the decoder both apply the 6-layer multi-head attention module. The teacher model adopts 3 self-attention layers. Similarly, the style discriminator uses 3 self-attention layers, while text discriminator takes one attention layer since it is adequate for judging whether a text is real or synthetic thanks to the large amount of word embedding parameters. 
We utilize the PyTorch 1.6 as our implementation framework. In this work, we build our model based the standard Transformer modules, and borrow several basic modules from OpenNMT-py.\footnote{https://github.com/OpenNMT/OpenNMT-py}
We use the Adam \cite{kingma2014adam} for optimization with $\beta 1$=0.5, $\beta 2$=0.98 and $\epsilon$ =$10^{-8}$ which works well with WGAN \cite{gulrajani2017improved}. We set the learning rate as 1e-4, and the dropout rate as 0.1. The probability of noising is set as 0.1. The $N_{rc}$, $N_{dr}$ and $N_{adr}$ in Algorithm~\ref{alg_training_proc} are 5, 1, 5 respectively. 
We train our model with GTX 1080Ti GPU. The maximum training step for Yelp is set to be 160000, and 160000 for IMDb, 320000 for Sentiment-Formality (multi-style transfer). The batch sizes are 96, 64 and 50 for Yelp, IMDb, and Sentiment-Formality, respectively. We save checkpoints and conduct validation every 5000, 5000, 10000 steps for these datasets respectively. We train our models three times with random seeds 123, 456, 789 independently, and then compute mean values and standard deviations.

we train baselines with NVIDIA A100, because some models of prior methods consume GPU memory more than 1080Ti's 11GB when they are trained with their best settings. We train these models three times independently, and also compute their mean values and standard deviations for fairness.

\begin{table}[t]
\vspace{-2ex}
    \centering
        \tabcolsep=4pt        \caption{\label{table_overall_performance} Automatic evaluation results against prior approaches on Yelp and IMDb. Numbers within parentheses are standard deviation. $^\dagger$Using the same data splits as ours, we evaluate the released transferred texts with our metrics because its source code fails to run due to missing required files. Under the dashline, we also present the results of ablation experiments. $\uparrow$ means the higher the score, the better the quality is, while $\downarrow$ means the smaller the better.}
    \scalebox{0.97} {
    \begin{tabular}{l c c c c}
        \toprule
        \textbf{Model} & \textbf{Acc}$\uparrow$ & \textbf{BLEU}$\uparrow$ & r-\textbf{BLEU}$\uparrow$ & \textbf{PPL}$\downarrow$    \\ \hline
        \multicolumn{5}{c}{\textbf{Yelp}} \\ \hline
    DelAndRet\shortcite{li2018delete} & 87.7\small{(0.8)} & 29.4\small{(2.2)} & 10.5\small{(1.5)} & 97.3\small{(4.6)}   \\
        MultiDecoder\shortcite{fu2018style} & 88.1\small{(4.6)} & 36.3\small{(3.8)} & 10.8\small{(0.9)} & 112.4\small{(12.1)}   \\
    StyTrans-cond\shortcite{dai-etal-2019-style}     &  91.5\small{(2.1)} & 41.6\small{(2.6)} & 18.3\small{(1.3)} & 98.5\small{(15.9)}  \\
        StyTrans-multi\shortcite{dai-etal-2019-style}    &  86.0\small{(2.2)} & 52.8\small{(3.5)} & 20.2\small{(1.5)} & 85.5\small{(13.8)}  \\
        DGST\shortcite{li2020dgst} & 86.6\small{(1.5)} & 54.1\small{(1.1)} & 20.8\small{(1.7)} & 85.8\small{(1.8)}  \\
        DIRR\shortcite{liu-etal-2021-learning}$^\dagger$ & 94.2\small{(N/A)} & 52.6\small{(N/A)} & 20.7\small{(N/A)} & 88.0\small{(N/A)}    \\
        BARTM\shortcite{lai2021generic} &   88.3\small{(1.3)} & 53.2\small{(5.3)} & 22.7\small{(1.8)} & 84.4\small{(3.7)}   \\
        \hdashline
        Our MSSRNet & \textbf{95.6\small{(0.3)}} & \textbf{58.8\small{(0.4)}} & \textbf{25.6\small{(0.3)}} & \textbf{67.2\small{(3.2)}} \\
        - w/o sequential &  92.7\small{(0.6)} & 54.3\small{(0.7)} & 23.1\small{(0.8)} & 78.1\small{(4.7)}   \\
        - w/o TSLearn    &  92.1\small{(0.8)} & 56.7\small{(1.1)} & 24.4\small{(1.2)} & 75.3\small{(3.3)}   \\
        \hline
        
        \multicolumn{5}{c}{\textbf{IMDb}} \\ \hline
        DelAndRet\shortcite{li2018delete} & 75.8\small{(3.3)} & 39.0\small{(2.1)} & N/A & 77.1\small{(3.2)} \\
            MultiDecoder\shortcite{fu2018style} & 87.3\small{(2.2)} & 39.4\small{(1.4)} & N/A & 86.3\small{(6.7)}   \\
        StyTrans-cond\shortcite{dai-etal-2019-style}     &  83.7\small{(3.2)} & 60.4\small{(4.1)} & N/A & 105.2\small{(13.3)}   \\
            StyTrans-multi\shortcite{dai-etal-2019-style}    &  78.9\small{(3.8)} & 64.3\small{(9.3)} & N/A & 107.4\small{(16.5)}   \\
        DGST\shortcite{li2020dgst} & 72.4\small{(2.2)} & 60.25\small{(4.3)} & N/A & 89.8\small{(17.1)}  \\
        DIRR\shortcite{liu-etal-2021-learning}$^\dagger$ & 82.7\small{(N/A)} & 64.2\small{(N/A)} & N/A & 83.1\small{(N/A)}  \\
        BARTM\shortcite{lai2021generic} & 82.6\small{(1.5)} & 60.2\small{(4.1)} & N/A & 81.6\small{(2.8)}   \\
    \hdashline
    Our MSSRNet &   \textbf{91.3\small{(0.2)}} & \textbf{64.7\small{(0.3)}} & N/A & \textbf{71.0\small{(1.7)}}  \\
        - w/o sequential & 90.1\small{(0.5)} & 60.4\small{(0.4)} & N/A & 81.6\small{(2.2)}  \\
        - w/o TSLearn    &  88.7\small{(0.8)} & 61.3\small{(0.6)} & N/A & 80.3\small{(2.1)} \\
        \bottomrule
    \end{tabular}
    }
    
    \vspace{-2ex}
\end{table}
\section{Results and Analyses}
For fairness, we measure and compare all methods with the uniform metrics. Considering the different data split, for those methods with released code, we train the original models with official parameter settings, and the results we obtain are comparable to those in the original papers, which shows that our replication is fair. By computing the mean values and standard deviations of three independent runs, the results are more robust and reliable. This approach ensures that the results are representative of the true performance of the models, and that the performance variance is reduced to a minimum.

\textbf{Overall performance} Table \ref{table_overall_performance} presents the comparison of the proposed method with existing state-of-the-art methods in the field of text style transfer, using the metrics of accuracy, BLEU \cite{papineni2002bleu}\footnote{using multi-bleu.perl script with default settings.}, and PPL. 
Compared with other methods, our model achieves the best results. 
The results show that most models perform well in content preservation, but at the cost of lower transfer accuracy, which indicates that it is easier for models to achieve high BLEU scores but lower transfer accuracy by simply copying the source text. However, the proposed method maintains a good balance between the two aspects, achieving a 91.3\% accuracy on the Imdb dataset with a clear improvement over prior methods. Additionally, the method generates more fluent texts compared to the other models, demonstrating the ability of the model to precisely control stylistic information through manipulating sequential style representation. Overall, the results indicate that the proposed method is able to foster a strong text style transfer model.

\begin{figure}[htp]
    \includegraphics[width=0.48 \textwidth]{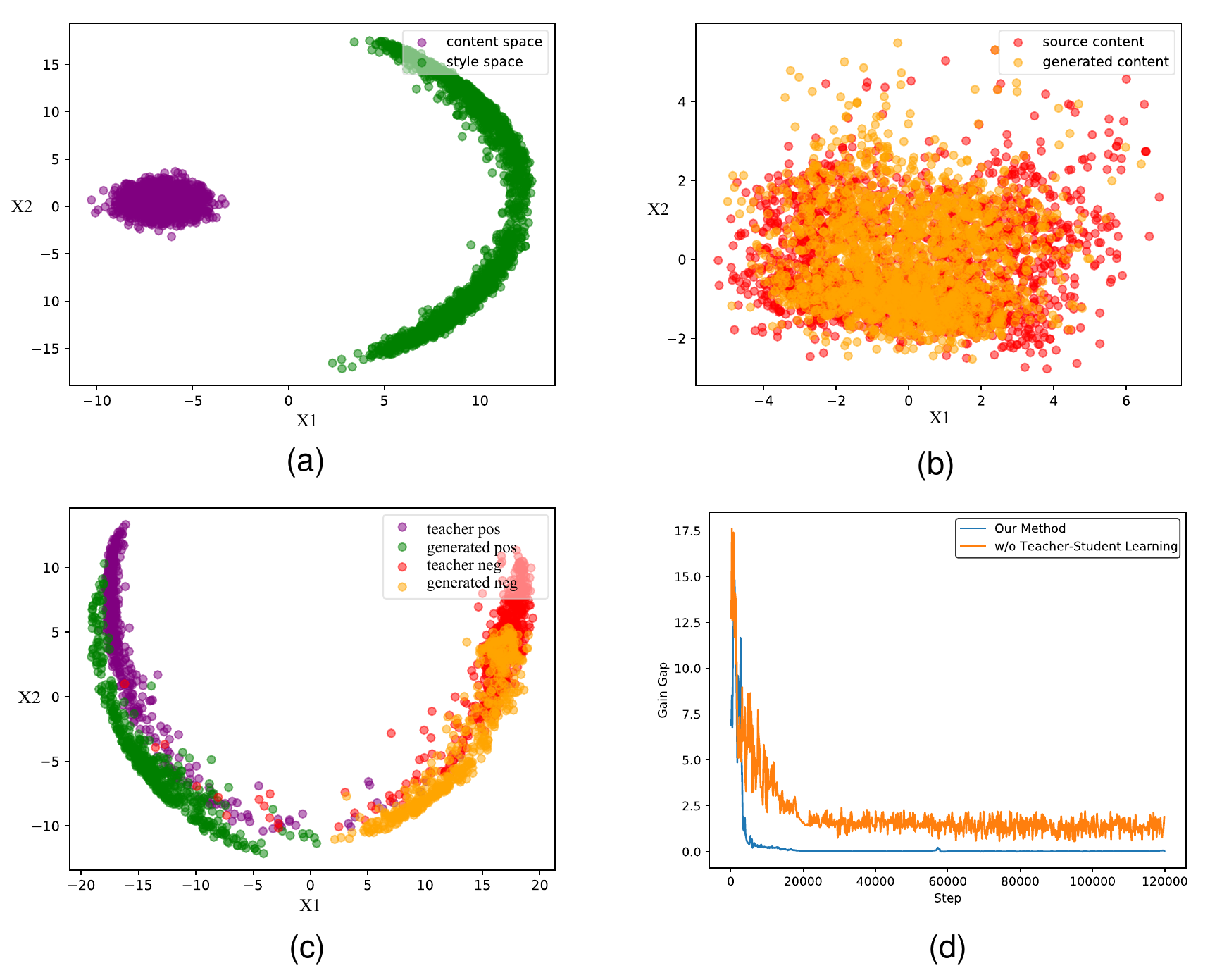}
    \caption{\label{fig_visualize} Visualization (a) over the content and style representation space. (b) over the content representation space of the source text and the transferred text. (c) over the style representation space of teacher and student model. (d) the gain gap between real samples and generated samples during training on Yelp. 
    }
    \vspace{-1.0ex}
\end{figure}

\textbf{Ablation Analysis} We replace the sequential style representation with fixed-sized vector representation. We see that its performance drops a lot on both datasets, which reflects that controlling sequential style representation is helpful to the performance improvement. Additionally, we remove the teacher-student learning strategy to see whether it is important to the performance. The results show that the performance decreases clearly. We also notice that the standard deviation values grow bigger when we remove the teacher-student learning, which indicates its contribution to the stable training. Meanwhile, models without using sequential style representation obtain higher PPL. This reflects that controlling fine-grained style strength has the potential to produce fluent sentences because a singular vector lacks of style discrimination among different words.

\textbf{Disentanglement \& Preservation} The t-SNE visualization is used to evaluate the learned representations. Subgraph (a) of Figure \ref{fig_visualize} reveals that the learned representations have separated content and style information into different spaces. This also indicates that the content representation is style-agnostic. Subgraph (b) of Figure \ref{fig_visualize} depicts the content representation space of the source text and that of the transferred text demonstrating. We find that these two content spaces overlap with each other closely. These findings support that our model is capable of generating the style-independent content representation and reserving its main content information.


\textbf{Teacher-Student Learning Analysis} The Style Generator (student) learns about how to produce sequential style representation from teacher model. We visualize the style representation generated by the teacher and student model. As shown in Subgraph (c) of Figure \ref{fig_visualize}, the representations produced by our model are close to those generated by teacher model, which proves that the style generator (student) effectively learns to produce sequential style representation similar to the teacher model. The positive and negative sentiment representations generated by the teacher model are well separated, which confirms the effectiveness of the teacher-student learning strategy in learning style representation.

\begin{figure}[htp]
    \includegraphics[width=0.48 \textwidth]{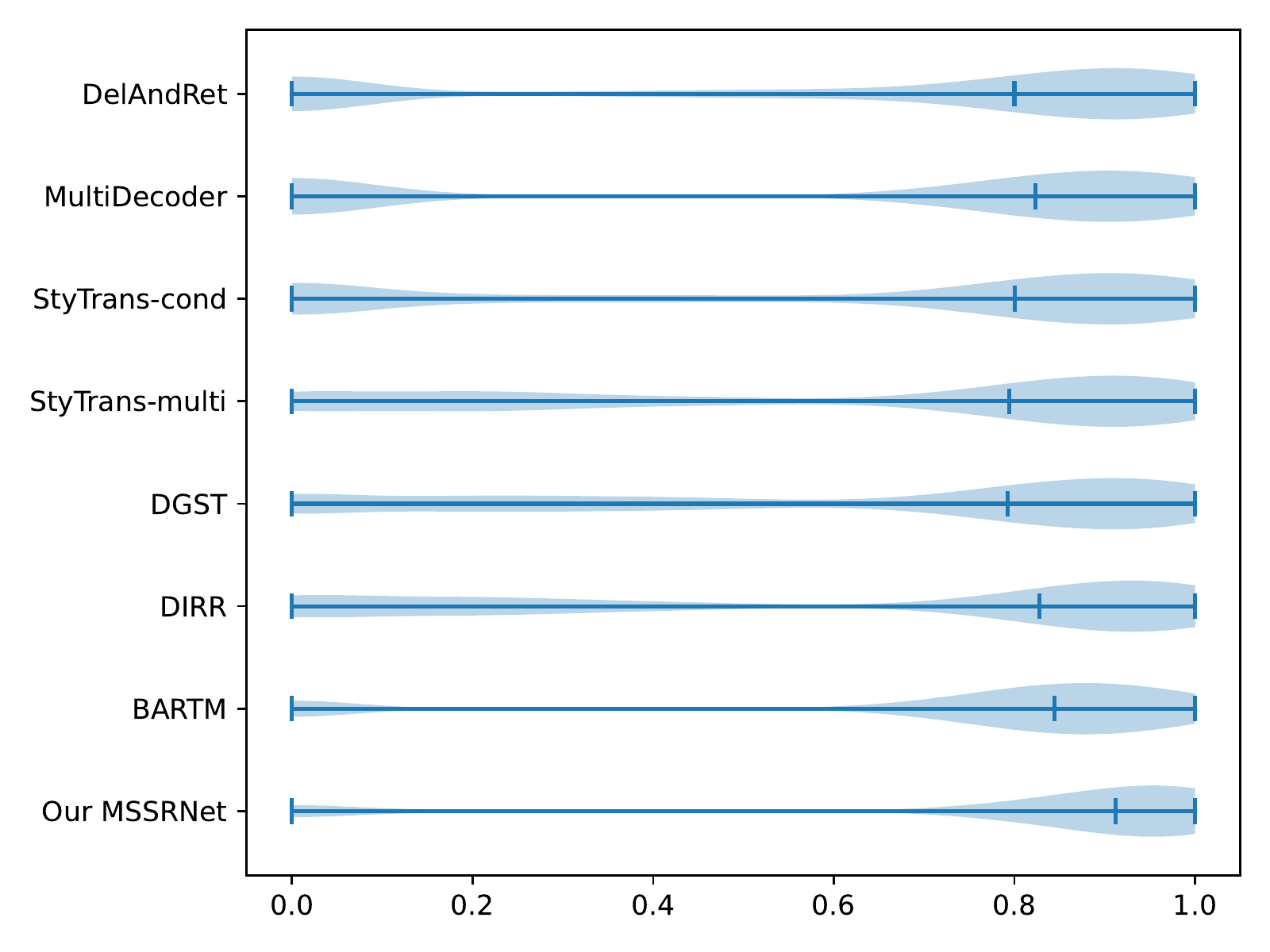}
    \caption{\label{fig_visualize_violin} Violin plot over the ratios of successfully converting stylistic tokens into target style on IMDb.
    }
\end{figure}
\textbf{Training Stability} 
Within the GAN framework, the training procedure goes with maximizing the gain~(feedback) to optimize the generator while minimizing the gain of fake example~(produced by generator) to optimize the discriminator. If the discriminator wins or loses by too big a margin, then the generator can’t learn as the feedback signal towards correct direction is too small. Hence, the gap between these two gains reflects the training stability~\cite{gulrajani2017improved}.
As we can see from the Subgraph (d) in Figure \ref{fig_visualize}, our method is much more stable with the help of teacher-student learning strategy. When we remove it, the gap shakes sharply as the training goes. 
Despite the benefit of applying sequential representation, manipulating such representation space directly is tough due to the difficulties of high dimension optimization.
However, when it learns from teacher model, we can optimize it towards stable direction by providing a supervised signal to the style generator and restricting the learned representation into a specific space. Thus it reduces the difficulties of adversarial training.

\textbf{Stylistic Tokens Transfer Proportion} The violin plot in Figure~\ref{fig_visualize_violin} displays the distribution of the proportion $r = num_s / num_a$, where $num_s$ represents the number of successfully transferred stylistic tokens in a text and $num_a$ is the total number of stylistic tokens in the text. The plot shows that our method is able to transfer a high proportion of stylistic tokens in each text, indicating its effectiveness in style transfer. The selection of stylistic tokens in each text is based on the explanation technology of the classifier~\cite{chen2016thorough,chen2018learn,ribeiro2016should,ribeiro2018anchors,li2016understanding,nguyen2018comparing,feng2018pathologies,gururangan2018annotation,thorne2019generating}, which involves finding the salient tokens for the current label by disturbing the classifier. The violin plot in Yelp and our selection details are shown in appendix, please refer to Appendix(\ref{appendix_style_toks_select}).
\begin{table}[htp]
    \centering
        \caption{\label{table_cases_selected_style_tokens} Examples of selected stylistic tokens.}
    \scalebox{0.95} {
    \begin{tabular}{p{1.0\columnwidth}}
        \toprule
        \multicolumn{1}{c}{\textbf{negative to positive}} \\ \hline
        \textbf{Text: } good food , great prices on wings on tuesdays . \\
            \textbf{Stylistic Tokens: } good, great \\
        \hdashline

            \textbf{Text: } our waitress was the best , very accommodating . \\
            \textbf{Stylistic Tokens: } best, accommodating \\
        \hdashline

            \textbf{Text: } this really is the worst film i have ever seen . \\
            \textbf{Stylistic Tokens: } worst \\
        \hdashline

            \textbf{Text: } all in all , madman is pretty much terrible and dull . \\
            \textbf{Stylistic Tokens: } much terrible, dull \\
            
        \bottomrule
    \end{tabular}
    }
\end{table}

\begin{figure}[htp]
    \includegraphics[width=0.48 \textwidth]{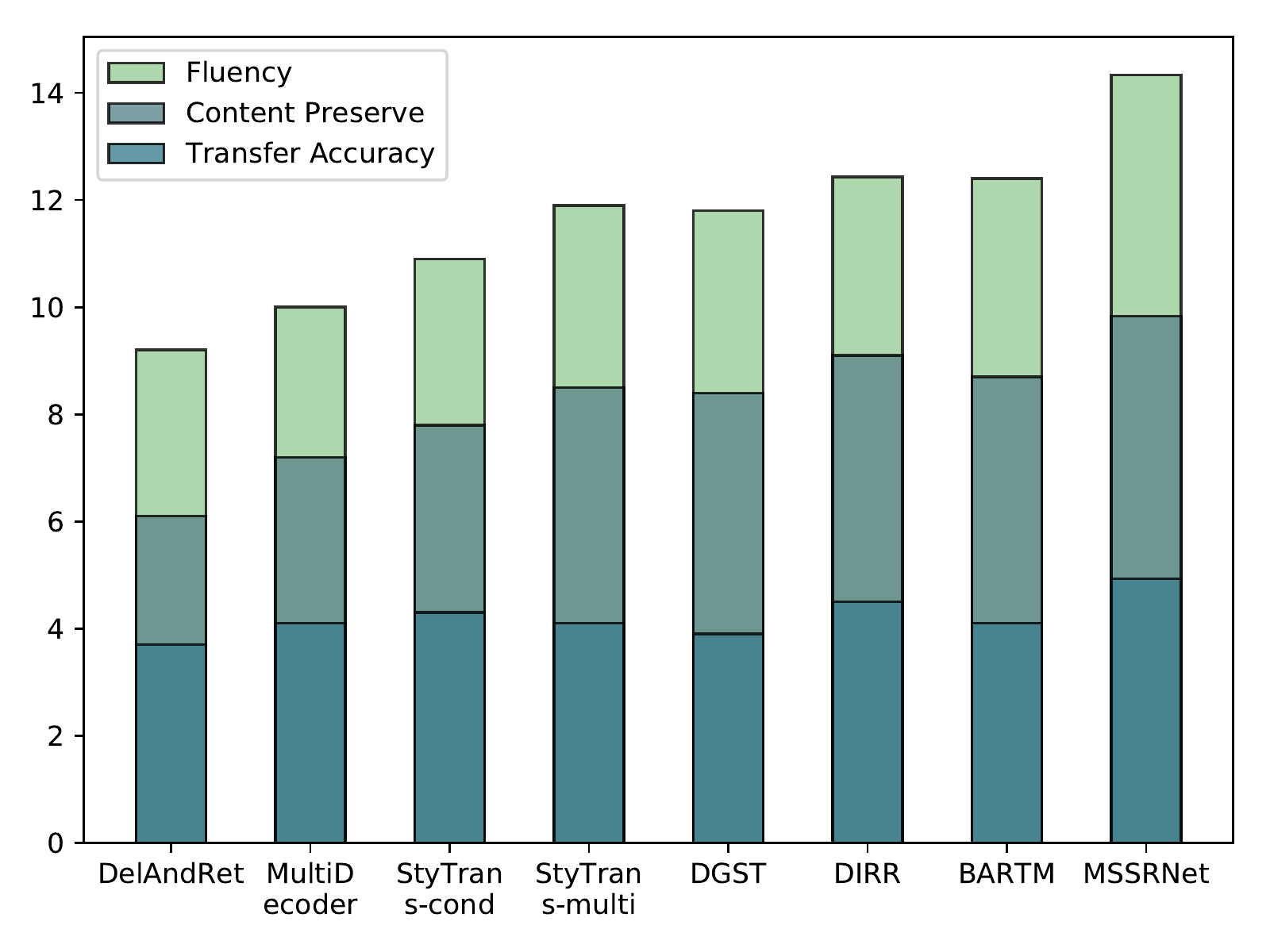}
    \caption{\label{fig_human_evaluation} Human evaluation in terms of fluency, content preservation and transfer accuracy.
    }
\end{figure}

\textbf{Human Evaluation} 
Except for the above automatic evaluations, we also conduct manual evaluation to further carry out comparison among different methods. We employed 10 human annotators to rate each sentence on a 0–5 scale (the higher the score, the better the quality is) in terms of
transfer accuracy, content preservation, and text fluency. We randomly selected 200 source sentences (100/100 sentences for positive/negative sentiment respectively) from the test set, and generated texts into target styles with different models. 
This evaluation was conducted in a strictly blind fashion: the transferred sentences generated by all compared methods were randomly shuffled, so that the annotator known nothing about which model produced a particular sentence. 
Figure~\ref{fig_human_evaluation} presents the comparison results which support the findings from the automatic evaluations, as they show that our method outperforms the prior approaches in every aspect.

\begin{table}[t]
    \centering
        \tabcolsep=4pt        \caption{\label{table_4styles_performance} Automatic evaluation results against prior approaches in multiple style transfer task. Numbers within parentheses are standard deviation.}
    \scalebox{0.95} {
    \begin{tabular}{l c c c c}
            \toprule
        \textbf{Model} & \textbf{Acc}$\uparrow$ &        \textbf{BLEU}$\uparrow$ & r-\textbf{BLEU}$\uparrow$ & \textbf{PPL}$\downarrow$ \\ 
            \hline
    MultiDecoder\shortcite{fu2018style} & 82.1\small{(0.7)} & 18.5\small{(2.1)} & 15.6\small{(0.5)} & 139.2\small{(7.3)}    \\
        StyTrans-multi\shortcite{dai-etal-2019-style}    &  81.3\small{(1.4)} & 51.2\small{(4.1)} & 31.9\small{(1.2)} & 154.2\small{(14.4)} \\
    Our MSSRNet & \textbf{89.9\small{(0.5)}} & \textbf{55.3\small{(0.5)}} & \textbf{32.5\small{(0.1)}} & \textbf{50.8\small{(1.2)}} \\
        \bottomrule
    \end{tabular}
    }
    
    \vspace{-2.0ex}
\end{table}

\textbf{Multi-Style Scenario}
We also conducted experiments on multi-style transfer dataset, and compared with previous methods that also support for multiple style transfer. Table~\ref{table_4styles_performance} presents the automatic evaluation results. Compared with StyTrans-multi~\cite{dai-etal-2019-style} that does not disentangle style and content representation, our method still copes well with transferring among multiple styles. This reveals that the disentangled representation and sequential style representation provide a more practical and efficient solution for multi-style transfer, leading to better results compared to previous methods. Besides, our method performs much better compared with MultiDecoder of which style information is represented with single vector. It further reinforces the effectiveness of manipulating sequential style representation.

\begin{table}[htp]
    \centering
        \caption{\label{table_cases_imdb} Transferred sentences generated by different models on IMDb test set. The words in {\color{blue}blue} and {\color{red}red} indicate good and bad transfer respectively.}
    \scalebox{0.8} {
    \begin{tabular}{p{1.27\columnwidth}}
        \toprule
        \multicolumn{1}{c}{\textbf{negative to positive}} \\ \hline
        \textbf{Source: } chevy chase is dull and amazingly unfunny . \\
        \hdashline
            \textbf{DelAndRet: } chevy chase {\color{red}and amazingly bad} . \\
            \textbf{MultiDecoder: } great do and is very funny and enjoy . \\
            \textbf{StyTrans-cond: } chevy chase is {\color{red}kavner and maneuvers} fantastic maneuvers mikkelsen . \\
            \textbf{StyTrans-multi: } chevy chase is intelligent and amazingly moving . \\
        \textbf{DGST: } chevy chase is {\color{red}scary} and amazingly {\color{blue}beautiful} . \\
        \textbf{DIRR: } chevy chase is {\color{blue}brilliant} and amazingly {\color{blue}funny} . \\
        \textbf{BARTM: } chevy chase is {\color{red}dull} and amazingly {\color{red}unfunny}. \\
        \textbf{Ours: } {\color{blue}definitely} chevy chase is {\color{blue}entertaining} and {\color{blue}amazingly hilarious} . \\ \hline
        \textbf{Source: } her performance is incredibly over-dramatic and overwrought in every way . \\
        \hdashline
            \textbf{DelAndRet: } her performance is incredibly {\color{red}annoying and unbelievable and} in every way . \\
            \textbf{MultiDecoder: } also the story is as good and in seeing it as . \\
            \textbf{StyTrans-cond: } her performance is incredibly {\color{red}maneuvers and maneuvers} in every way . \\
            \textbf{StyTrans-multi: } her performance is incredibly {\color{red}over-dramatic} and ironic in every way . \\
        \textbf{DGST: } her work is very {\color{red}irresistible} and affecting in every way . \\
        \textbf{DIRR: } her performance is incredibly {\color{blue}well} and {\color{red}overwrought} in every way . \\
        \textbf{BARTM: } her performance is incredibly good and overwrought in every way. \\
        \textbf{Ours: } her performance is incredibly {\color{blue}wonderful} and {\color{blue}superb} in every way . \\ \hline
        \multicolumn{1}{c}{\textbf{positive to negative}} \\ \hline
        \textbf{Source: } the peoples court is definitely the best of the tv court shows . \\
        \hdashline
            \textbf{DelAndRet: } the peoples court is definitely the tv court shows . \\
            \textbf{MultiDecoder: } the only saving grace is this one in the characters ever made . \\
            \textbf{StyTrans-cond: } the peoples court is definitely the {\color{blue}worst} of the tv court shows . \\
            \textbf{StyTrans-multi: } the peoples court is {\color{blue}not the best} of the tv court shows . \\
        \textbf{DGST: } the peoples court is {\color{red}barely a actors of the the western town} . \\
        \textbf{DIRR: } the peoples court is definitely the {\color{blue}worst} of the tv court shows . \\
        \textbf{BARTM: } the peoples court is definitely the {\color{blue}worst} of the tv court shows . \\
        \textbf{Ours: } the peoples court is definitely the {\color{blue}worst} of the tv court shows . \\ \hline
        \textbf{Source: } i have a deep love for and passion about movies like this one . \\
        \hdashline
            \textbf{DelAndRet: } i have a deep {\color{red}love for romance} and about everything about this movie or any other comedy . \\
            \textbf{MultiDecoder: } i have been an insult to bad movies and look like this movie : . \\
            \textbf{StyTrans-cond: } i have a {\color{red}joke love for and passion} about movies like this one . \\
            \textbf{StyTrans-multi: } i have a {\color{red}deep love} for movies like this one . \\
        \textbf{DGST: } i have a {\color{blue}little} love for and passion {\color{red}all} movies like this one . \\
        \textbf{DIRR: } i have a {\color{red}shallow bad} for and passion about movies like this one . \\
        \textbf{BARTM: } i have a deep {\color{red}love and passion} for movies like this . \\
        \textbf{Ours:} i have a deep {\color{blue}disappointment} about movies like this one . \\ 
        \bottomrule
    \end{tabular}
    }
\end{table}
\begin{table}[htp]
    \centering
        \caption{\label{table_cases_4class} Transferred sentences generated by different models on multi-style test set. The words in {\color{blue}blue} and {\color{red}red} indicate good and bad transfer respectively.}
    \scalebox{0.86} {
    \begin{tabular}{p{1.15\columnwidth}}
        \toprule
        \textbf{Source: } the food here is bland and boring and bad . \\
            \hdashline
            \multicolumn{1}{c}{\textbf{negative to positive}} \\
        \hdashline
            \textbf{MultiDecoder: } the food here is {\color{blue}delicious} and reasonable happy ... friendly . \\
            \textbf{StyTrans-multi: } the food here is {\color{blue}tasteful and lively and fantastic} . \\
        \textbf{Ours: } the food here is {\color{blue}delicious and interesting and good} . \\ 
            \hdashline
            \multicolumn{1}{c}{\textbf{negative to negative \& informal}} \\
            \hdashline
            \textbf{MultiDecoder: } {\color{blue}btw} the food i made ... and very {\color{red}great food . !} \\
            \textbf{StyTrans-multi: } {\color{red}the movie} that is bland and boring and bad . \\
        \textbf{Ours: } {\color{blue}horrible ! ! !} the food here is really bland and boring and bad ! \\
            \hdashline
            \multicolumn{1}{c}{\textbf{negative to negative \& formal}} \\
            \hdashline
            \textbf{MultiDecoder: } the food {\color{red}was the man and too good man .} \\
            \textbf{StyTrans-multi: } {\color{red}the movie answer} is rejected and boring and pleasing . \\
        \textbf{Ours: } {\color{blue}i think} the food here is bland , boring and bad . \\
            \hline
            
            \textbf{Source: } but it is definitely worth the wait . \\
            \hdashline
            \multicolumn{1}{c}{\textbf{positive to negative}} \\
        \hdashline
            \textbf{MultiDecoder: } but it is do {\color{blue}not} watch it ? \\
            \textbf{StyTrans-multi: } but it is {\color{blue}never} {\color{red}waiting} the wait . \\
        \textbf{Ours: } but it is {\color{blue}not worth} the wait . \\ 
            \hdashline
            \multicolumn{1}{c}{\textbf{positive to positive \& informal}} \\
            \hdashline
            \textbf{MultiDecoder: } but it is do be very sure . \\
            \textbf{StyTrans-multi: } but it is definitely worth the wait . \\
        \textbf{Ours: } it is definitely worth the wait {\color{blue}! :)} \\
            \hdashline
            \multicolumn{1}{c}{\textbf{positive to positive \& formal}} \\
            \hdashline
            \textbf{MultiDecoder: } but it is definitely be good it ? \\
            \textbf{StyTrans-multi: } {\color{red}they} it is definitely worth the wait . \\
        \textbf{Ours: } my answer is it is definitely worth the wait . \\
            \hline
        
            \textbf{Source: } check out this link ... lots of places \& prices ! \\
            \hdashline
            \multicolumn{1}{c}{\textbf{informal to informal \& negative}} \\
        \hdashline
            \textbf{MultiDecoder: } {\color{red}this place another place for taco desk , and coffee !} \\
            \textbf{StyTrans-multi: } check out this {\color{red}rating} ... lots of places \& prices ! \\
        \textbf{Ours: } {\color{blue}check out this elsewhere , there are nothing .} \\ 
            \hdashline
            \multicolumn{1}{c}{\textbf{informal to informal \& positive}} \\
            \hdashline
            \textbf{MultiDecoder: } check this place {\color{red}off far , large , and coffee !} \\
            \textbf{StyTrans-multi: } check out this link ... lots of places \& prices ! \\
        \textbf{Ours: } check out this link . lots of places \& prices {\color{blue}will be found !} \\
            \hdashline
            \multicolumn{1}{c}{\textbf{informal to formal}} \\
            \hdashline
            \textbf{MultiDecoder: } check out {\color{red}answer up songs or agree , and video .} \\
            \textbf{StyTrans-multi: } check out this link . \\
        \textbf{Ours: } {\color{blue}please} check out this link , {\color{blue}there are} lots of places and prices . \\
            \hline

            \textbf{Source: } that is a good start . \\
            \hdashline
            \multicolumn{1}{c}{\textbf{formal to formal \& negative}} \\
        \hdashline
            \textbf{MultiDecoder: } that is a {\color{red}good} start . \\
            \textbf{StyTrans-multi: } that is a rd start . \\
        \textbf{Ours: } that is a {\color{blue}bad} start . \\ 
            \hdashline
            \multicolumn{1}{c}{\textbf{formal to formal \& positive}} \\
            \hdashline
            \textbf{MultiDecoder: } that is a {\color{blue}good} visit ! \\
            \textbf{StyTrans-multi: } {\color{blue}that is a good start .} \\
        \textbf{Ours: } {\color{blue}that is a good start .} \\
            \hdashline
            \multicolumn{1}{c}{\textbf{formal to informal}} \\
            \hdashline
            \textbf{MultiDecoder: } that is a good goes 2 . \\
            \textbf{StyTrans-multi: } that is a good start . \\
        \textbf{Ours: } {\color{blue}hey , it would be} a good start . \\
            
        \bottomrule
    \end{tabular}
    }
    
\end{table}

\textbf{Case Study}
Table \ref{table_cases_imdb} displays some comparisons of transferred texts generated by different models on test set of IMDb. 
We observe that most of the models preserve the main content of the source text.
For the first example, DIIR and our MSSRNet transfer all stylistic words in a suitable manner, while others fail to achieve it. We also notice that prior methods may remain some words unchanged which convey source style when there are more than one parts needing to be changed. In contrast, our model not only preserves the content of the text, but also successfully changes multiple style-relevant words to the target style, resulting in a better overall transfer performance. This highlights the advantage of our approach in terms of preserving content and accurately converting the style of the text. Cases on test set of Yelp are shown in Table~\ref{table_cases_yelp}. We observe that most models perform well when the text is easy to transfer. But when it encounters with complex texts, such as the second example in Table~\ref{table_cases_yelp}, prior methods fail to convert all stylistic phrases correctly. 
More cases of multi-style transfer task are shown in Table~\ref{table_cases_4class}. Similarly, we can observe that our method still alters text style better under the scenario of multiple style transfer. For the second example, we find that our method successfully converts the text style into target one, that is positive and informal at the same time, through adding a smiling face symbol.

\section{Conclusions}
In conclusion, this research paper presents a novel approach to address the challenges encountered in text style transfer by manipulating sequential style representation instead of fixed-sized vector representation. The proposed method integrates a teacher-student learning strategy within an adversarial training framework, enabling direct optimization signal and mitigating training complexity. Extensive experiments and analyses are conducted to comprehensively evaluate the effectiveness of the proposed method. The experimental results demonstrate superior performance in terms of transfer accuracy, content preservation, and linguistic fluency, particularly in multi-style transfer scenarios. Future investigations will focus on exploring the application of this method to unsupervised or semi-supervised Neural Machine Translation (NMT) tasks. Since the objective of translation involves preserving the meaning of a text while converting it from one language to another, which bears resemblance to preserving the content of a text while altering its style, the fine-grained style transfer method proposed in this paper holds promising potential for adaptation to language translation tasks.

\bibliographystyle{ACM-Reference-Format}
\bibliography{mssrnet}

\appendix
\label{sec:appendix}

\section{Appendix}
\subsection{Stylistic Tokens Selection}
\label{appendix_style_toks_select}
A line of approaches~\cite{chen2016thorough,ribeiro2018anchors,li2016understanding, nguyen2018comparing,feng2018pathologies,gururangan2018annotation,thorne2019generating} towards explainable text classifier is to highlight the important parts of inputs to explain the predictions of neural networks. Actually, they usually disturb each token of the input and determine whether it is salient token according to the change of prediction result. Similarly, we rely on disturbing the weight of each token to select vital tokens. Firstly, we build and train a text classifier for each style transfer dataset. Our classifier is:
\begin{equation}
\label{eq_word_select_eq1}
\begin{aligned}
\textbf{\emph{S}}_{cls} = {\rm Enc} (Emb(X))
\end{aligned}
\end{equation}
where $\textbf{\emph{S}}_{cls} = \{ \mathbf{s}_1, \mathbf{s}_2, ..., \mathbf{s}_n \} $ is the sequence of style vectors for tokens from $\rm X$, n is the length of $\rm X$.

\begin{equation}
\label{eq_word_select_eq2}
\begin{aligned}
\alpha_i = \frac{{\rm exp}(\mathbf{v}_{cls}^{\top} \mathbf{s}_i)}{\sum_{i'}{\rm exp}(\mathbf{v}_1^{\top} \mathbf{s}_{i'})} \\
\textbf{\emph{v}} = \sum_i \alpha_i \mathbf{s}_i
\end{aligned}
\end{equation}
\begin{equation}
\label{eq_word_select_eq3}
\begin{aligned}
\textbf{\emph{d}}_{cls} = softmax( \textbf{W}_{cls}^{\rm T} \textbf{\emph{v}} + \textbf{b}_{cls} )
\end{aligned}
\end{equation}
where $\mathbf{v}_{cls}$, $\textbf{W}_{cls}$ and $\textbf{b}_{cls}$ are trainable parameters, $\textbf{\emph{d}}_{cls} \in \mathbb{R}^K $, $K$ is the number of styles.

\begin{figure}[!htbp]
    \includegraphics[width=0.42 \textwidth]{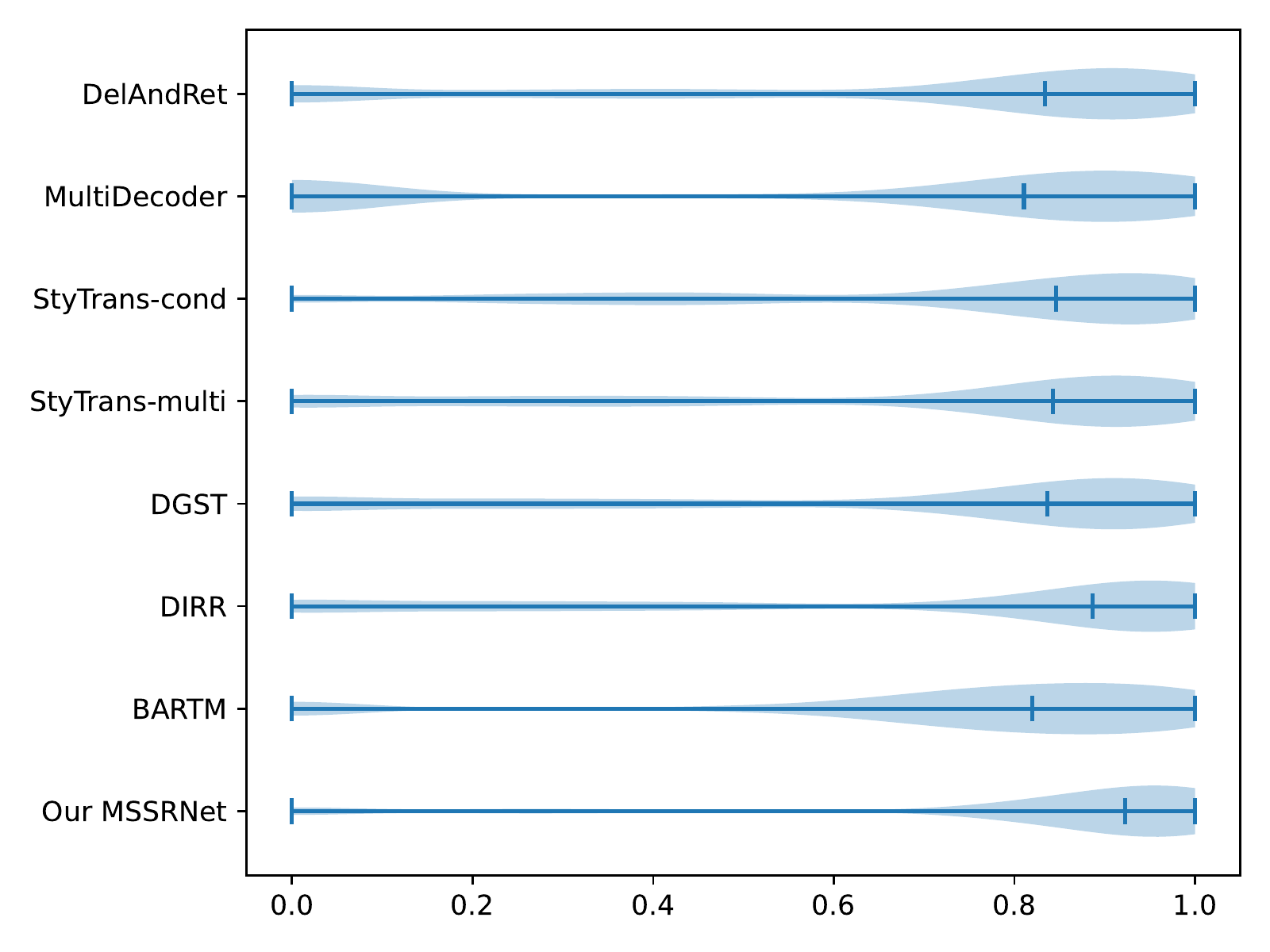}
    \caption{\label{fig_visualize_violin_yelp} Violin plot over the ratios of successfully converting stylistic tokens into target style on Yelp.
    }
\vspace{-2.0ex}
\end{figure}
\textbf{Our disturb method} We adopt the token span (k-gram) as disturbing unit so that we may select important phrases not just single token, and finally we use 1-gram, 2-gram and 3-gram. While disturbing, we set $\alpha_i = 0.0$ for each token \textit{i} in the span. We select those token spans whose disturbances lead to the drop of prediction probability bigger than threshold $\beta$. In our experiments, we find that setting $\beta$ to be 10\% works well. Table~\ref{table_cases_selected_style_tokens} shows some examples including selected stylistic tokens. 
The violin plot on the test set of Yelp is shown in Figure~\ref{fig_visualize_violin_yelp}. Similarly to that in IMDb, our method performs better against prior methods.

\subsection{Case Study Examples}

\textbf{Case Study on Yelp}
Table~\ref{table_cases_yelp} shows some examples that generated by different methods on the test set of Yelp.

\begin{table}[htbp]
    \centering
        \caption{\label{table_cases_yelp} Transferred sentences generated by different models on Yelp test set. The words in {\color{blue}blue} and {\color{red}red} indicate good and bad transfer respectively.}
    \scalebox{0.92} {
    \begin{tabular}{p{1.1\columnwidth}}
        \toprule
        \multicolumn{1}{c}{\textbf{negative to positive}} \\ \hline
        \textbf{Source: } the food was pretty bad , i would not go there again . \\
        \hdashline
            \textbf{DelAndRet: } the food was definitely , i will definitely come back there again . \\
            \textbf{MultiDecoder: } the food was pretty {\color{blue}good} , is i highly recommended is it ! \\
            \textbf{StyTrans-cond: } the food is pretty {\color{blue}good} , i would {\color{blue}definitely go} there again . \\
            \textbf{StyTrans-multi: } the food , i would {\color{red}not go} there need . \\
        \textbf{DGST: } the food was pretty {\color{blue}good} , i would {\color{blue}definitely go} there again . \\
        \textbf{DIRR: } the food was pretty {\color{blue}good} , i {\color{blue}will go} there again . \\
        \textbf{BARTM: } the food was pretty {\color{blue}good} , i would {\color{blue}go} there again . \\
        \textbf{Ours: } the food was pretty {\color{blue}delicious} , i would {\color{blue}recommend} go there again . \\ \hline
        \textbf{Source: } there are no smiles and no customer service . \\
        \hdashline
            \textbf{DelAndRet: } there are also good and service . \\
            \textbf{MultiDecoder: } {\color{red}there are no toast , many customer service .} \\
            \textbf{StyTrans-cond: } {\color{red}there are to doug and available s. service .} \\
            \textbf{StyTrans-multi: } there are customer service . \\
        \textbf{DGST: } there are great smiles and {\color{blue}great} customer service . \\
        \textbf{DIRR: } there are great smiles and great \\
        \textbf{BARTM: } this is no budget film-making at its very, very best. \\
        \textbf{Ours: } there are {\color{blue}warm} smiles and {\color{blue}good} customer service . \\ \hline
        \multicolumn{1}{c}{\textbf{positive to negative}} \\ \hline
        \textbf{Source: } great place to hang out , grab a meal and a few brews ! \\
        \hdashline
            \textbf{DelAndRet: } {\color{red}place out} , i grab a meal and a little lunch {\color{red}even left} ! \\
            \textbf{MultiDecoder: } {\color{red}terrible is sorry , me come dinner and a friend service before .} \\
            \textbf{StyTrans-cond: } {\color{red}\$ \_num\_ to i out} , grab a meal and a few brews ? \\
            \textbf{StyTrans-multi: } \_num\_ minutes grab meal brews ! \\
        \textbf{DGST: } {\color{red}\$ \_num\_ to ran out} , grab a meal and a few establishments ! \\
        \textbf{DIRR: } {\color{blue}terrible} place to hang out , grab a meal and a few brews ! \\
        \textbf{BARTM: } {\color{blue}bad} place to hang out , grab a meal and a few brews ! \\
        \textbf{Ours: } {\color{blue}horrible} place to hang out , grab a meal and a few brews ! \\ \hline
        \textbf{Source: } the atmosphere was fun and the staff treats you well . \\
        \hdashline
            \textbf{DelAndRet: } and the staff treats {\color{blue}terribly} . \\
            \textbf{MultiDecoder: } the atmosphere was {\color{red}good , the same and it really .} \\
            \textbf{StyTrans-cond: } the time was i discounted the staff nephew you management . \\
            \textbf{StyTrans-multi: } the atmosphere caf staff service- you . \\
        \textbf{DGST: } the atmosphere was nothing and the staff treats bother nothing . \\
        \textbf{DIRR: } the atmosphere was {\color{blue}gross} and the staff treats you {\color{blue}poorly} .  \\
        \textbf{BARTM: } the atmosphere was {\color{blue}bad} and the staff treats you {\color{blue}badly} . \\
        \textbf{Ours:} the atmosphere was {\color{blue}not fun} and the staff treats you {\color{blue}rudely} . \\ 
        \bottomrule
    \end{tabular}
    }
    
\end{table}


\end{document}